# A Critical Assessment of
# Benchmark Comparison in Planning

**Adele E. Howe**                                    HOWE@CS.COLOSTATE.EDU
**Eric Dahlman**                                    DAHLMAN@CS.COLOSTATE.EDU
*Computer Science Department*
*Colorado State University, Fort Collins, CO 80523*

## Abstract

Recent trends in planning research have led to empirical comparison becoming commonplace. The field has started to settle into a methodology for such comparisons, which for obvious practical reasons requires running a subset of planners on a subset of problems. In this paper, we characterize the methodology and examine eight implicit assumptions about the problems, planners and metrics used in many of these comparisons. The problem assumptions are: PR1) the performance of a general purpose planner should not be penalized/biased if executed on a sampling of problems and domains, PR2) minor syntactic differences in representation do not affect performance, and PR3) problems should be solvable by STRIPS capable planners unless they require ADL. The planner assumptions are: PL1) the latest version of a planner is the best one to use, PL2) default parameter settings approximate good performance, and PL3) time cut-offs do not unduly bias outcome. The metrics assumptions are: M1) performance degrades similarly for each planner when run on degraded runtime environments (e.g., machine platform) and M2) the number of plan steps distinguishes performance. We find that most of these assumptions are not supported empirically; in particular, that planners are affected differently by these assumptions. We conclude with a call to the community to devote research resources to improving the state of the practice and especially to enhancing the available benchmark problems.

## 1. Introduction

In recent years, comparative evaluation has become increasingly common for demonstrating the capabilities of new planners. Planners are now being directly compared on the same problems taken from a set of domains. As a result, recent advances in planning have translated to dramatic increases in the size of the problems that can be solved (Weld, 1999), and empirical comparison has highlighted those improvements.

Comparative evaluation in planning has been significantly influenced and expedited by the Artificial Intelligence Planning and Scheduling (AIPS) conference competitions. These competitions have had the dual effect of highlighting progress in the field and providing a relatively unbiased comparison of state-of-the-art planners. When individual researchers compare their planners to others, they include fewer other planners and fewer test problems because of time constraints.

To support the first competition in 1998 (McDermott, 2000), Drew McDermott defined, with contributions from the organizing committee, a shared problem/domain definition language, PDDL (McDermott et al., 1998) (Planning Domain Definition Language). Using a common language means that planners' performance can be directly compared, without entailing hand translation or factoring in different representational capabilities.

As a second benefit, the lack of translation (or at least human accomplished translation) meant that performance could be compared on a large number of problems and domains[1]. In fact, the five competition planners were given a large number of problems (170 problems for the ADL track and 165 for the STRIPS track) within seven domains, including one domain that the planner developers had never seen prior to the competition. So the first competition generated a large collection of benchmarks: seven domains used in the competition plus 21 more that were considered for use. All 28 domains are available at ftp://ftp.cs.yale.edu/pub/mcdermott/domains/. The second competition added three novel domains to that set.

A third major benefit of the competitions is that they appear to have motivated researchers to develop systems that others can use. The number of entrants went from five in the first competition to 16 in the second. Additionally, all of the 1998 competitors and six out of sixteen of the 2000 competitors made their code available on web sites. Thus, others can perform their own comparisons.

In this paper, we describe the current practice of comparative evaluation as it has evolved since the AIPS competitions and critically examine *some* of the underlying assumptions of that practice. We summarize existing evidence about the assumptions and describe experimental tests of others that had not previously been considered. The assumptions are organized into three groups concerning critical decisions in the experiment design: the problems tested, the planners included and the performance metrics collected.

Comparisons (as part of competitions or by specific researchers) have proven to be enormously useful to motivating progress in the field. Our goal is to understand the assumptions so that readers know how far the comparative results can be generalized. In contrast to the competitions, the community cannot legislate fairness in individual researcher's comparative evaluations, but readers may be able to identify cases in which results should be viewed either skeptically or with confidence. Thus, we conclude the paper with some observations and a call for considerably more research into new problems, metrics and methodologies to support planner evaluation.

Also in contrast to the competitions, our goal is *not* to declare a winner. Our goal is also *not* to critique individual studies. Consequently, to draw attention away from such a possible interpretation, whenever possible, we report all results using letter designators that were assigned randomly to the planners.

## 2. Planning Competitions and Other Direct Comparisons

Recently, the AIPS competitions have spurred considerable interest in comparative evaluation. The roots of comparative planner evaluation go back considerably further, however. Although few researchers were able to run side-by-side comparisons of their planners with others, they were able to demonstrate performance of their planner on well-known problems, which could be viewed as de facto benchmarks. Sussman's anomaly (Sussman, 1973) in Blocksworld was the premier planning benchmark problem and domain for many years; every planner needed to "cut its teeth" on it.

---

1. To solve a particular planning problem (i.e., construct a sequence of actions to transform an initial state to a goal state), planners require a domain theory and a problem description. The domain theory represents the abstract actions that can be executed in the environment; typically, the domain descriptions include variables that can be instantiated to specific objects or values. Multiple problems can be defined for each domain; problem descriptions require an initial state description, a goal state and an association with some domain.

As researchers tired of Blocksworld, many called for additional benchmark problems and environments. Mark Drummond, Leslie Kaelbling and Stanley Rosenschein organized a workshop on benchmarks and metrics (Drummond, Kaelbling, & Rosenschein, 1990). Testbed environments, such as Martha Pollack's TileWorld (Pollack & Ringuette, 1990) or Steve Hanks's TruckWorld (Hanks, Nguyen, & Thomas, 1993), were used for comparing algorithms within planners. By 1992, UCPOP (Penberthy & Weld, 1992) was distributed with a large set of problems (117 problems in 21 domains) for demonstration purposes. In 1995, Barry Fox and Mark Ringer set up a planning and scheduling benchmarks web page (`http://www.newosoft.com/~benchmrx/`) to collect problem definitions, with an emphasis on manufacturing applications. Recently, PLANET (a coordinating organization for European planning and scheduling researchers) has proposed a planning benchmark collection initiative (http://planet.dfki.de).

Clearly, benchmark problems have become well-established means for demonstrating planner performance. However, the practice has known benefits and pitfalls; Hanks, Pollack and Cohen (1994) discuss them in some detail in the context of agent architecture design. The benefits include providing metrics for comparison and supporting experimental control. The pitfalls include a lack of generality in the results and a potential for the benchmarks to unduly influence the next generation of solutions. In other words, researchers will construct solutions to excel on the benchmarks, regardless of whether the benchmarks accurately represent desired real applications.

To obtain the benefits just listed for benchmarks, the problems often are idealized or simplified versions of real problems. As Cohen (1991) points out , most research papers in AI, or at least at an AAAI conference, exploit benchmark problems; yet few of them relate the benchmarks to target tasks. This may be a significant problem; for example, in a study of flowshop scheduling[2] benchmarks, we found that performance on the standard benchmark set did not generalize to performance on problems with realistic structure (Watson, Barbulescu, Howe, & Whitley, 1999). A study of just Blocksworld problems found that the best known Blocksworld benchmark problems are atypical in that they require only short plans for solution and optimal solutions are easy to find (Slaney & Thiebaux, 2001).

In spite of these difficulties, benchmark problems and the AIPS competitions have considerably influenced comparative planner evaluations. For example, in the AIPS 2000 conference proceedings (Chien, Kambhampati, & Knoblock, 2000), all of the papers on improvements to classical planning (12 out of 44 papers at the conference) relied heavily on comparative evaluation using benchmark problems; the other papers concerned scheduling, specific applications, theoretical analyses or special extensions to the standard paradigm (e.g., POMDP, sensing). Of the 12 classical papers, six used problems from the AIPS98 competition benchmark set, six used problems from Kautz and Selman's distribution of problems with blackbox (Kautz, 2002) and three added some of their own problems as well. Each paper showed results on a subset of problems from the benchmark distributions (e.g., Drew McDermott's from the first competition) with logistics, blocksworld, rocket and gripper domains being most popular (used in 11, 7, 5 and 5 papers, respectively). The availability of planners from the competition was also exploited; eight of the papers compared their systems to other AIPS98 planners: blackbox, STAN, IPP and HSP (in 5, 3, 3 and 1 papers, respectively).

---

[2]. Scheduling is an area related to planning in which the actions are already known, but their *sequence* still needs to be determined. Flowshop scheduling is a type of manufacturing scheduling problem.

## 3. Assumptions of Direct Comparison

A canonical planner evaluation experiment follows the procedure in Table 1. The procedure is designed to compare performance of a new planner to the previous state of the art and highlight superior performance in some set of cases for the new planner. The exact form of an experiment depends on its purpose, e.g., showing superiority on a class of problem or highlighting the effect of some design decision.

---

1. Select and/or construct a subset of planner domains
2. Construct problem set by:
   - running large set of benchmark problems
   - selecting problems with desirable features
   - varying some facet of the problem to increase difficulty (e.g., number of blocks)
3. Select other planners that are:
   - representative of the state of the art on the problems OR
   - similar to or distinct from the new planner, depending on the point of the comparison or advance of the new planner OR
   - available and able to parse the problems
4. Run all problems on all planners using default parameters and setting an upper limit on time allowed
5. Record which problems were solved, how many plan steps/actions were in the solution and how much CPU time was required to either solve the problem, fail or time out

---

Table 1: Canonical comparative planner evaluation experiment.

The protocol depends on three selections: problems, planners and evaluation metrics. It is simply not practical or even desirable to run all available planners on all available problems. Thus, one needs to make informed decisions about which to select. A purpose of this paper is to examine the assumptions underlying these decisions to help make them more informed. Every planner comparison does not adopt every one of these assumptions, but the assumptions are ones commonly found in planner comparisons. For example, those comparisons designed for a specific purpose (e.g., to show scale-up on certain problems or suitability of the planner for logistics problems) will carefully select particular types of problems from the benchmark sets.

**Problems**   Many planning systems were developed to solve a particular type of planning problem or explore a specific type of algorithmic variation. Consequently, one would expect them to perform better on the problems on which and for which they were developed. Even were they not designed for a specific purpose, the test set used during development may have subtly biased the development. The community knows that planner performance depends on problem features, but not in general, how, when and why. Researchers tend to design planners to be general purpose. Consequently, comparisons assume that

> *the performance of a general-purpose planner should not be penalized/biased if executed on a sampling of problems and domains* (problem assumption 1).

The community also knows that problem representation influences planner performance. For example, benchmark problem sets include many versions of Blocksworld problems, designed by different planner developers. These versions vary in their problem representation, both minor apparently syntactic changes (e.g., how clauses are ordered within operators, initial conditions and goals, and whether any information is extraneous) and changes reflecting addition of domain knowledge (e.g., what constraints are included and whether variables are typed). Consequently, comparisons assume that

> *syntactic representational modifications either do not matter or affect each planner equally* (problem assumption 2).

PDDL includes a field, `:requirements`, for the capabilities required of a planner to solve the problem. PDDL1.0 defined 21 values for the `:requirements` field; the base/default requirement is `:strips`, meaning STRIPS derived add and delete sets for action effects. `:adl` (from Pednault's Action Description Language) requires variable typing, disjunctive preconditions, equality as a built-in predicate, quantified preconditions and conditional effects in addition the `:strips` capability. Yet, many planners either ignore the `:requirements` field or reject the problem only if it specifies `:adl` (ignoring many of the other requirements that could also cause trouble). Thus, comparisons assume that

> *problems in the benchmark set should be solvable by a STRIPS planner unless they require* `:adl` (problem assumption 3).

**Planners**   The wonderful trend of making planners publicly available has led to a dilemma in determining which to use and how to configure them. The problem is compounded by the longevity of some of these planner projects; some projects have produced multiple versions. Consequently, comparisons tend to assume that

> *the latest version of the planner is the best* (planner assumption 1).

These planners may also include parameters. For example, the blackbox planner allows the user to define a strategy for applying different solution methods. Researchers expect that parameters affect performance. Consequently, comparisons assume that

> *default parameter settings approximate good performance* (planner assumption 2).

Experiments invariably use time cut-offs for concluding planning that has not yet found a solution or declared failure. Many planners would need to exhaustively search a large space to declare failure. For practical reasons, a time out threshold is set to determine when to halt a planner, with a failure declared when the time-out is reached. Thus, comparisons assume that

> *if one picks a sufficiently high time-out threshold, then it is highly unlikely that a solution would have been found had slightly more time been granted* (planner assumption 3).

**Metrics**  Ideally, performance would be measured based on how well the planner does its job (i.e., constructing the 'best' possible plan to solve the problem) and how efficiently it does so. Because no planner has been shown to solve all possible problems, the basic metric for performance is the number or percentage of problems actually solved within the allowed time. This metric is commonly reported in the competitions. However, research papers tend not to report it directly because they typically test a relatively small number of problems.

Efficiency is clearly a function of memory and effort. Memory size is limited by the hardware. Effort is measured as CPU time, preferably but not always on the same platform in the same language. The problems with CPU time are well known: programmer skill varies; research code is designed more for fast prototyping than fast execution; numbers in the literature cannot be compared to newer numbers due to processor speed improvements. However, if CPU times are regenerated in the experimenter's environment then one assumes that

> *performance degrades similarly with reductions in capabilities of the runtime environment (e.g., CPU speed, memory size)* (metric assumption 1).

In other words, an experimenter or user of the system does not expect that code has been optimized for a particular compiler/operating system/hardware configuration, but it should perform similarly when moved to another compatible environment.

The most commonly reported comparison metric is computation time. The second most is number of steps or actions (for planners that allow parallel execution) in a plan. Although planning seeks solutions to achieving goals, the goals are defined in terms of states of the world, which does not lend itself well to general measures of quality. In fact, quality is likely to be problem dependent (e.g., resource cost, amount of time to execute, robustness), which is why number of plan steps has been favored. Comparisons assume that

> *number of steps in a resulting plan varies between planner solutions and approximates quality* (metric assumption 2).

Any comparison, competitions especially, has the unenviable task of determining how to trade-off or combine the three metrics (number solved, time, and number of steps). Thus, if number of steps does not matter, then the comparison could be simplified.

We converted each assumption into a testable question. We then either summarized the literature on the question or ran an experiment to test it.

## 3.1 Our Experimental Setup

Some of the key issues have been examined previously, directly or indirectly. For those, we simply summarize the results in the subsections that follow. However, some are open questions. For those, we ran seven well known planners on a large set of 2057 benchmark problems. The planners all accept the PDDL representation, although some have built-in translators for PDDL to their internal representation and others rely on translators that we added. When several versions of a planner were available, we included them all (for a total of 13 planners). The basic problem set comprises the UCPOP benchmarks, the AIPS98 and 2000 competition test sets and an additional problem set developed for a specific application.

With the exception of the permuted problems (see the section on Problem Assumption 2 for specifics), the problems were run on 440 MHz Ultrasparc 10s with 256 Megabytes of memory running SunOS 2.8. Whenever possible, versions compiled by the developers were used; when only source code was available, we compiled the systems according to the developers' instructions. The planners written in Common Lisp were run under Allegro Common Lisp version 5.0.1. The other planners were compiled with GCC (EGCS version 2.91.66). Each planner was given a 30 minute limit of wall clock time[3] to find a solution; however, all times reported are run times returned by the operating system.

### 3.1.1 PLANNERS

The planners are all what have been called *primitive-action planners* (Wilkins & desJardins, 2001), planners that require relatively limited domain knowledge and construct plans from simple action descriptions. Because the AIPS98 competition required planners to accept PDDL, the majority of planners used in this study were competition entrants or are later versions thereof [4]. The common language facilitated comparison between the planners without having to address the effects of a translation step. The two exceptions were UCPOP and Prodigy; however, their representations are similar to PDDL and were translated automatically. The planners represent five different approaches to planning: plan graph analysis, planning as satisfiability, planning as heuristic search, state-space planning with learning and partial order planning. When possible, we used multiple versions of a planner, and not necessarily the most recent. Because we conducted this study over some period of time (almost 1.5 years), we froze the set early on; we are not comparing the performance to declare a winner and so did not think that the lack of recent versions undermined the results of testing our assumptions.

**IPP** (Koehler, Nebel, Hoffmann, & Dimopoulos, 1997) extends the Graphplan (Blum & Furst, 1997) algorithm to accept a richer plan description language. In its early versions, this language was a subset of ADL that extends the STRIPS formalism of Graphplan to allow for conditional and universally quantified effects in operators. Until version 4.0, negation was handled via the introduction of new predicates for the negated preconditions and corresponding mutual exclusion rules; subsequent versions handle it directly (Koehler, 1999). We used the AIPS98 version of IPP as well as the later 4.0 version.

---

3. We used actual time on lightly loaded machines because occasionally a system would thrash due to inadequate memory resulting in little progress over considerable time.
4. We used the BUS system as the manager for running the planners (Howe, Dahlman, Hansen, Scheetz, & von Mayrhauser, 1999), which was implemented with the AIPS98 competition planners. This facilitated the running of so many different planners, but did somewhat bias what was included.

**SGP**   (Sensory Graph Plan) (Weld, Anderson, & Smith, 1998) also extends Graphplan to a richer domain description language, primarily focusing on uncertainty and sensing. As with IPP, some of this transformation is performed using expansion techniques to remove quantification. SGP also *directly* supports negated preconditions and conditional effects. SGP tends to be slower (it is implemented in Common Lisp instead of C) than some of the other Graphplan based planners. We used SGP version 1.0b.

**STAN**   (STate ANalysis) (Fox & Long, 1999) extends the Graphplan algorithm in part by adding a preprocessor (called *TIM*) to infer type information about the problem and domain. This information is then used within the planning algorithm to reduce the size of the search space that the Graphplan algorithm would search. STAN also incorporated optimized data structures (bit vectors of the planning graph) that help avoid many of the redundant calculations performed by Graphplan. Additionally, STAN maintains a *wave front* during graph construction to track remaining goals and so limit graph construction. Subsequent versions incorporated further analyses (e.g., symmetry exploitation) and an additional simpler planning engine. Four versions of STAN were tested: the AIPS98 competition version, version 3.0, version 3.0s and a development snapshot of version 4.0.

**blackbox**   (Kautz & Selman, 1998) converts planning problems into Boolean satisfiability problems, which are then solved using a variety of different techniques. The user indicates which techniques should be tried in what order. In constructing the satisfiability problem, blackbox uses the planning graph constructed as in Graphplan. For blackbox, we used version 2.5 and version 3.6b.

**HSP**   (Heuristic Search Planner) (Bonet & Geffner, 1999) is based on heuristic search. The planner uses a variation of hill-climbing with random restarts to solve planning problems. The heuristic is based on using the Graphplan algorithm to solve a relaxed form of the planning problem. In this study, we used version 1.1, which is an algorithmic refinement of the version entered into the AIPS98 competition, and version 2.0.

**Prodigy**   [5] (The Prodigy Research Group, 1992) combines state-space planning with backward chaining from the goal state. A plan under construction consists of a head-plan of totally ordered actions starting from the initial state and a tail-plan of partially ordered actions related to the goal state. Although not officially entered into the competition, informal results presented at the AIPS98 competition suggested that Prodigy performed well in comparison to the entrants. We used Prodigy version 4.0.

**UCPOP**   (Barrett, Golden, Penberthy, & Weld, 1993) is a Partial Order Causal Link planner. The decision to include UCPOP was based on several factors. First, it does not expand quantifiers and negated preconditions; for some domains, the expansion from grounding operators can be so great as to make the problem insolvable. Second, UCPOP is based on a significantly different algorithm in which interest has recently resurfaced. We used UCPOP version 4.1.

---

5. We thank Eugene Fink for code that translates PDDL to Prodigy.

| Source | # of Domains | # of Problems |
|---|---|---|
| Benchmarks | 50 | 293 |
| AIPS 1998 | 6 | 202 |
| AIPS 2000 | 5 | 892 |
| Developers | 1 | 13 |
| Application | 3 | 72 |

Table 2: Summary of problems in our testing set: source of the problems, the number of domains and problems within those domains.

### 3.1.2 TEST PROBLEMS

Following standard practice, our experiments require planners to solve commonly available benchmark problems and the AIPS competition problems. In addition, to test our assumptions about the influence of domains (assumption PR1) and representations of problems (assumption PR2), we will also include permuted benchmark problems and some other application problems. This section describes the set of problems and domains in our study, focusing on their source and composition.

The problems require only STRIPS capabilities (i.e., add and delete lists). We chose this least common denominator for several reasons. First, more capable planners can still handle STRIPS requirements; thus, this maximized the number of planners that could be included in our experiment. Also, not surprisingly, more problems of this type are available. Second, we are examining assumptions of evaluation, including the effect of required capabilities on performance. We do not propose to duplicate the effort of the competitions in singling out planners for distinction, but rather, our purpose is to determine what factors differentially affect planners.

The bulk of the problems came from the AIPS98 and AIPS 2000 problem sets and the set of problems distributed with the PDDL specification. The remaining problems were solicited from several sources. The source and counts of problems and domains are summarized in Table 2.

**Benchmark Problems** The preponderance of problems in planning test sets are "toy problems": well-known synthetic problems designed to test some attribute of planners. The Blocksworld domain has long been included in any evaluation because it is well known, can have subgoal interactions and supports constructing increasingly complex problems (e.g., towers of more blocks). A few benchmark problems are simplified versions of realistic planning problems, e.g., the flat tire, refrigerator repair or logistics domains. We used the set included with the UCPOP planner. These problems were contributed by a large number of people and include multiple encodings of some problems/domains, especially Blocksworld.

**AIPS Competitions: 1998 and 2000** For the first AIPS competition, Drew McDermott solicited problems from the competitors as well as constructing some of his own, such as the mystery domain, which had semantically useless names for objects and operators. Problems were generated for each domain automatically. The competition included 155 problems from six domains: robot movement in a grid, gripper in which balls had to be moved between rooms by a robot with two grippers, logistics of transporting packages, organizing snacks for movie watching, and two mystery domains, which were disguised logistics problems.

The format of the 1998 competition required entrants to execute 140 problems in the first round. Of these problems, 52 could not be solved by any planner. For round two, the planners executed 15 new problems in three domains, one of which had not been included in the first round.

The 2000 competition attracted 15 competitors in three tracks: STRIPS, ADL and a hand-tailored track. It required performance on problems in five domains: logistics, Blocksworld, parts machining, Freecell (a card game), and Miconic-10 elevator control. These domains were determined by the organizing committee, with Fahiem Bacchus as the chair, and represented a somewhat broader range. We chose problems from the Untyped STRIPS track for our set.

From a scientific standpoint, one of the most interesting conclusions of both competitions was the observed trade-offs in performance. Planners appeared to excel on different problems, either solving more from a set or finding a solution faster. In 1998, IPP solved more problems and found shorter plans in round two; STAN solved its problems the fastest; HSP solved the most problems in round one; and blackbox solved its problems the fastest in round one. In 2000, awards were given to two groups of distinguished planners across the different categories of planners (STRIPS, ADL and hand tailored), because according to the judges, "it was impossible to say that any one planner was the best"(Bacchus, 2000); TalPlanner and FF were in the highest distinguished planner group. The graphs of performance do show differences in computation time relative to other planners and to problem scale-up. However, each planner failed to solve some problems, which makes these trends harder to interpret (the computation time graphs have gaps).

The purpose of these competitions was to showcase planner technology at which they succeeded admirably. The planners solved much harder problems than could have been accomplished in years past. Because of this trend in planners handling increasingly difficult problems, the competition test sets may become of historical interest for tracking the field's progress.

**Problems Solicited from Planner Developers**   We also asked planner developers what problems she had used during development. One developer, Maria Fox, sent us a domain (Sodor, which is a logistics application) and set of problems that they had used. We would have included other domains and problems had we received any others.

**Other Applications**   The Miconic elevator domain from the AIPS2000 competition was derived from an actual planning application. The domain and problems were extremely simplified (e.g., removing the arithmetic).

To add another realistic problem to the comparison, we included one other planning application to the set of test domains: generating cases to test a software interface. Because of the similarities between software interface test cases and plans, we developed a system, several years ago, for automatically generating interface test cases using an AI planner. The system was designed to generate test cases for the user interface to Storage Technology's robot tape library (Howe, von Mayrhauser, & Mraz, 1997). The interface (i.e., the commands in the interface) was coded as the domain theory. For example, the *mount* com-mand/action's description required that a drive be empty and had the effect of changing the position of the tape being mounted and changing the status of the tape drive. Problems described initial states of the tape library (e.g., where tapes were resident, what was the status of the devices and software controller) and goal states that a human operator might wish to achieve.

At the time, we found that only the simplest problems could be generated using the planners available. We included this application in part because we knew it would be a challenge. As part of the test set, we include three domain theories (different ways of coding the application involving 8-11 operators) and twenty-four problems for each domain. We included only 24 because we wanted to include enough problems to see some effect, but not too many to overly bias the results. These problems were relatively simple, requiring the movement of no more than one tape coupled with some status changes, but they were still more difficult than could be solved in our original system.

## 3.2 Problem Assumptions

General-purpose planners exhibit differential capabilities on domains and sometimes even problems within a domain. Thus, the selection of problem set would seem to be critical to evaluation. For example, many problems in benchmark sets are variants of logistics problems; thus, a general-purpose planner that was actually tailored for logistics may appear to be better overall on current benchmarks. In this section, we will empirically examine some possible problem set factors that may influence performance results.

**Problem Assumption 1: To What Extent Is Performance of General Purpose Planners Biased Toward Particular Problems/Domains?** Although most planners are developed as general purpose, the competitions and previous studies have shown that planners excel on different domains/problems. Unfortunately, the community does not yet have a good understanding of why a planner does well on a particular domain. We studied the impact of problem selection on performance in two ways.

First, we assessed whether performance might be positively biased toward problems tested during development. Each developer[6] was asked to indicate which domains they used during development. We then compared each planner's performance on their development problems (i.e., the development set) to the problems remaining in the complete test set (rest). We ran 2x2 $\chi^2$ tests comparing number of problems solved versus failed in the development and test sets. We included only the number solved and failed in the analysis as timed-out problems made no difference to the results[7].

The results of this analysis are summarized in Table 3; Figure 1 graphically displays the ratio of successes to failures for the development and other problems. All of the planners except C performed significantly better on their development problems. This suggests that these planners have been tailored (intentionally or not) for particular types of problems and that they will tend to do better on test sets biased accordingly. For example, one of the planners in our set, STAN, was designed with an emphasis on logistics problems (Fox & Long, 1999).

---

6. We decided against studying some of the planners in this way because the representations for their development problems were not PDDL.

7. One planner was the exception to this rule; in one case, the planner timed out far more frequently on non-development problems.

| Planner | Development | | Rest | | $\chi^2$ | P |
|---|---|---|---|---|---|---|
| | Sol. | Fail | Sol. | Fail | | |
| A | 48 | 56 | 207 | 1026 | 51.70 | 0.001 |
| B | 42 | 34 | 226 | 929 | 51.27 | 0.001 |
| C | 30 | 0 | 549 | 16 | 0.13 | 0.722 |
| G | 43 | 35 | 233 | 924 | 49.56 | 0.001 |
| H | 52 | 9 | 234 | 655 | 91.41 | 0.001 |
| I | 113 | 20 | 328 | 920 | 187.72 | 0.001 |
| J | 114 | 24 | 388 | 949 | 157.62 | 0.001 |
| K | 37 | 56 | 203 | 987 | 27.82 | 0.001 |
| L | 63 | 32 | 358 | 846 | 52.13 | 0.001 |

Table 3: $\chi^2$ results comparing outcome on development versus other problems.

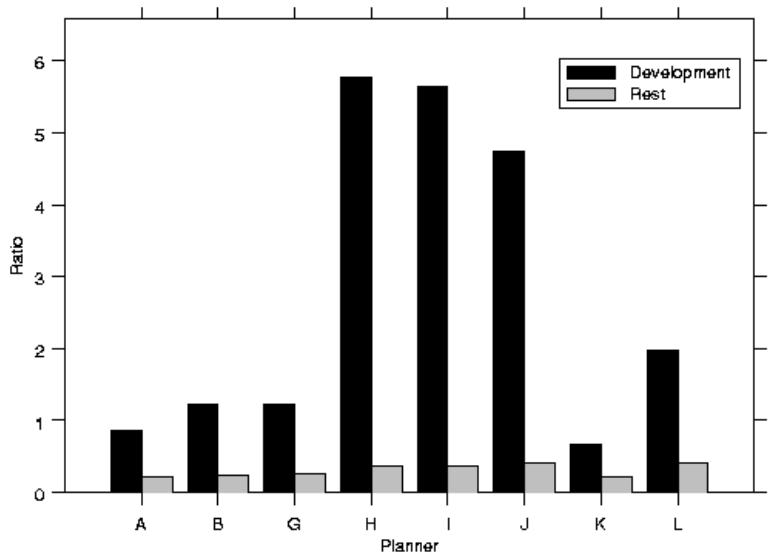

Figure 1: Histogram of ratios of success/failures for development and other problems for each of the planners.

The above analysis introduces a variety of biases. The developers tended to give us short lists that probably were not really representative of what they actually used. The set used is a moving target, rather than stationary as this suggests. The set of problems included in experimentation for publication may be different still. Consequently, for the second part, we broadened the question to determine the effect of different subsets of problems on performance. For each of 10 trials, we randomly selected $n$ domains (and their companion problems) to form the problem set. We counted how many of these problems could be solved by each planner and then ranked the relative performance of each planner. Thus, for each value of $n$, we obtained 10 planner rankings. We focused on rankings of problems solved for two reasons: First, each domain includes a different number of problems, making the count of problems variable across each of the trials. Second, relative ranking gets to the heart of whether one planner might be considered to be an improvement over another.

| | Rank Dominance | | | | | | | | | | | Total Pairs |
|---|---|---|---|---|---|---|---|---|---|---|---|---|
| $n$ | 0 | 1 | 2 | 3 | 4 | 5 | 6 | 7 | 8 | 9 | 10 | |
| 5 | 0 | 1 | 2 | 0 | 5 | 7 | 10 | 4 | 10 | 18 | 21 | 78 |
| 10 | 0 | 3 | 0 | 0 | 4 | 10 | 6 | 7 | 5 | 23 | 20 | 78 |
| 20 | 0 | 0 | 0 | 0 | 1 | 3 | 8 | 7 | 11 | 8 | 40 | 78 |
| 30 | 0 | 0 | 0 | 0 | 1 | 1 | 9 | 6 | 9 | 8 | 44 | 78 |

Table 4: Rank dominance counts for 10 samples of domains with domain sizes ($n$) of five through 30.

We tested values of 5, 10, 20 and 30 for $n$ (30 is half of the domains at our disposal). To give a sense of the variability in size, at $n = 5$, the most problems solved in a trial varied from 11 to 64. To assess the changes in rankings across the trials, we computed *rank dominance* for all pairs of planners; rank dominance is defined as the number of trials in which planner $x$'s rank was lower than planner $y$'s (note: ties would count toward neither planner). The 13 planners in our study resulted in 78 dominance pairings. If the relative ranking between two planners is stable, then one would expect one to always dominate the other, i.e., have rank dominance of 10.

Table 4 shows the number of pairs having each value (0-10) of rank dominance for the four values for $n$. For a given pair, we used the highest number as the rank dominance for the pair, e.g., if one always has a lower rank, then the pair's rank dominance is 10 or if both have five, then it is five. Because of ties, the maximum can be less than five. The data suggest that even when picking half of the domains, the rankings are not completely stable: in 56% of the pairings, one always dominates, but 22% have a 0.3 or greater chance of switching relative ranking. The values degrade as $n$ decreases with only 27% always dominating for $n = 5$.

**Problem Assumption 2: How Do Syntactic Representation Differences Affect Performance?**  Although it is well known that some planners' performance depends on representation (Joslin & Pollack, 1994; Srinivasan & Howe, 1995), two recent developments in planner research suggest that the effect needs to be better understood. First, a common representation, i.e., PDDL, may bias performance. Some planners rely on a pre-processing step to convert PDDL to their native representation, a step that usually requires making arbitrary choices about ordering and coding. Second, an advantage of planners based on Graphplan is that they are supposed to be less vulnerable to minor changes in representa-tion. Although the reasoning for the claim is sound, the exigencies of implementation may require re-introduction of representation sensitivity.

| Planner | All | None | Subset |
|---------|-----|------|--------|
| A | 65 | 315 | 30 |
| B | 70 | 295 | 45 |
| C | 318 | 74 | 18 |
| D | 202 | 169 | 39 |
| E | 111 | 132 | 167 |
| F | 112 | 138 | 160 |
| G | 70 | 295 | 45 |
| H | 91 | 290 | 29 |
| I | 109 | 134 | 167 |
| J | 150 | 124 | 136 |
| K | 60 | 305 | 45 |
| L | 112 | 284 | 14 |
| M | 212 | 148 | 50 |

Table 5: The number of problems for which the planners were able to solve all, none or only a subset of the permutations.

To evaluate the sensitivity to representation, ten permutations of each problem in the AIPS2000 set were generated, resulting in 4510 permuted problems. The permutations were constructed by randomly reordering the preconditions in the operator definitions and the order of the definitions of the operators within the domain definition.

We limited the number of problems in this study because ten permutations of all problems would be prohibitive. We selected the AIPS2000 problems for attention because this was the most recently developed benchmark set. Even within that set, not all of the domains were permuted because some would not result in different domains under the transformation we used. For the purposes of this investigation, we limited the set of modifications to permutations of preconditions and operators because these were known to affect some planners and because practical considerations limited the number of permutations that could be executed. Finally, for expediency, we ran the permutations on a smaller number of faster platforms because it expedited throughput and computation time was not a factor in this study.

To analyze the data, we divided the performance on the permutations of the problems into three groups based on whether the planner was able to solve all of the permutations, none of the permutations or only a subset of the permutations. If a planner is insensitive to the minor representational changes, then the subset count should be zero. From the results in Table 5, we can see that all of the planners were affected by the permutation operation. The susceptibility to permuting the problem was strongly planner dependent ($\chi^2 = 1572.16$, $P < 0.0001$), demonstrating that some planners are more vulnerable than others.

By examining the number in the Subset column, one can assess the degree of susceptibility. All of the planners were sensitive to reorderings, even those that relied on Graphplan methodology. The most sensitive were E, F, I and J (which included some Graphplan based planners and in which 40% of the problems had mixed results on the permutations) with C and L being least sensitive (3-4% were affected).

| Planner | Feature | | | | | | | | |
|---------|--------|------------|-----------|----------|--------|--------|--------|--------|---------|
|         | Axioms | Cond. Eff. | Dis. Pre. | Equality | ∃ Pre. | Safety | Strips | Typing | ∀ Pre. |
| A | 0 | 0 | 0 | 35 | 0 | 0 | 255 | 8 | 0 |
| B | 0 | 0 | 0 | 8 | 0 | 0 | 268 | 0 | 0 |
| C | 5 | 169 | 165 | 216 | 163 | 2 | 561 | 197 | 160 |
| D | 3 | 164 | 166 | 196 | 139 | 0 | 279 | 180 | 139 |
| E | 1 | 162 | 152 | 199 | 157 | 0 | 384 | 168 | 149 |
| F | 0 | 157 | 145 | 185 | 150 | 0 | 376 | 165 | 145 |
| G | 0 | 0 | 0 | 8 | 0 | 0 | 276 | 0 | 0 |
| H | 0 | 0 | 0 | 46 | 0 | 0 | 285 | 17 | 0 |
| I | 0 | 138 | 138 | 169 | 138 | 0 | 441 | 139 | 138 |
| J | 0 | 130 | 130 | 160 | 130 | 0 | 502 | 130 | 130 |
| K | 0 | 0 | 0 | 8 | 0 | 0 | 240 | 0 | 0 |
| L | 0 | 19 | 13 | 24 | 16 | 0 | 421 | 13 | 13 |
| M | 0 | 168 | 169 | 212 | 149 | 2 | 372 | 180 | 151 |

Table 6: The number of problems claiming to require each PDDL feature solved by each planner.

**Problem Assumption 3: Does Performance Depend on PDDL Requirements Features?** The planners were all intended to handle STRIPS problems. Some of the problems in the test set claim to require features other than STRIPS; one would expect that some of the planners would not be able to handle those problems. In addition, those planners that claim to be able to handle a given feature may not do as well as other planners. Table 6 shows the effects of feature requirements on the ability to solve problems. The data in this table are based on the features specified with the `:requirements` list in the PDDL definition of the domain.

We did not verify that the requirements were accurate or necessary; thus, the problem may be solvable by ignoring a part of the PDDL syntax that is not understood, or the problem may have been mislabeled by its designer. This is evident in cases where a planner that does not support a given feature still appears to be able to solve the corresponding problem. Some planners, e.g., older versions of STAN, will reject any problem that requires more than STRIPS without trying to solve it; an ADL problem that only makes use of STRIPS features would not be attempted.

As guidance on which planner to use when, these results must be viewed with some skepticism. For example, it would appear based on these results that planner I might be a good choice for problems with conditional effects as it was able to solve many of these problems. This would be a mistake, since that planner cannot actually handle these types of problems. In these cases, the problems claim to require ADL, but in fact, they only make use of the STRIPS subset.

Clearly, certain problems can only be solved by specific planners. For instance, C and M are the only planners that are able to handle safety constraints, while based on the data, only C, D and E appear to handle domain axioms. About half the planners had trouble with the typed problems. Some of the gaps appear to be due to problems in the translation to native representation.

## 3.3 Planners

Publicly available, general-purpose planners tend to be large programs developed over a period of years and enhanced to include additional features over time. Thus, several versions are likely to be available, and those versions are likely to have features that can be turned on/off via parameter settings.

When authors release later versions of their planning systems, the general assumption is that these newer versions will outperform their predecessors. However, this may not be the case in practice. For instance, a planner could be better optimized toward a specific class of problem which then in turn hurts its performance on other problems. Also, advanced capabilities, even when unused, may incur overhead in the solution of all problems.

So for comparison purposes, should one use the latest version? First, we tested this question in a study comparing multiple versions of four of the planners. Second, each planner relies on parameter settings to tune its performance. Some, such as blackbox, have many parameters. Others have none. Comparisons tend to use the default or published parameter settings because few people usually understand the effects of the parameters and tuning can be extremely time consuming. So does this practice undermine a fair comparison?

**Planner Assumption 1: Is the Latest Version the Best?** In this study, we compared performance of multiple versions of four planners (labeled for this section with W, X, Y and Z, with larger version numbers indicating subsequent versions). We considered two criteria for improvement: outcome of planning and computation time for solved problems. The outcome of planning is one of: solved, failed or timed-out. On each criterion, we statistically analyzed the data for superior performance of one of the versions. The outcome results for all the planners are summarized in Table 7. As the table shows, rarely does a new version result in more problems being solved. Only Z improved the number of our test problems solved in subsequent versions.

To check for whether the differences in outcome are significant, we ran 2x3 $\chi^2$ tests with planner version as independent variable and outcome as dependent. Table 8 summarizes the results of the $\chi^2$ analysis. For Z, we compared each version to its successor only. The differences are significant except for Y and the transition from Z 2 to 3 (this was expected because these two versions were extremely similar).

Another planner performance metric, which we evaluated, was the speed of solution. For this analysis, we limited the comparison to just those problems that were solved by both versions of the planner. We then classified each problem by whether the later version solved the problem faster, slower, or in the same time as the preceding version. From the results in Table 9, we see that all of the planners improved in the average speed of solution for subsequent versions, with the exception of Z (transition from the 1 to 2 versions). However, Z did increase the number of problems solved between those versions.

| Planner | Version | Solved | Failed | Timeout | Δ Solved? |
|---------|---------|--------|--------|---------|-----------|
| W | 1 | 286 | 664 | 533 | |
| W | 2 | 255 | 1082 | 147 | ⇓ |
| X | 1 | 502 | 973 | 3 | |
| X | 2 | 441 | 940 | 103 | ⇓ |
| Y | 1 | 387 | 750 | 339 | |
| Y | 2 | 382 | 771 | 329 | ⇓ |
| Z | 1 | 240 | 1043 | 201 | |
| Z | 2 | 276 | 959 | 248 | ⇑ |
| Z | 3 | 268 | 963 | 252 | ⇓ |
| Z | 4 | 421 | 878 | 184 | ⇑ |

Table 7: Version performance: counts of outcome and change in number solved.

| Planner | old Version | new Version | $\chi^2$ | P |
|---------|-------------|-------------|----------|---|
| W | 1 | 2 | 320.96 | .0001 |
| X | 1 | 2 | 98.84 | .0001 |
| Y | 1 | 2 | .46 | .79 |
| Z | 1 | 2 | 10.96 | .004 |
| Z | 2 | 3 | .158 | .924 |
| Z | 3 | 4 | 48.50 | .0001 |

Table 8: $\chi^2$ results comparing versions of the same planner.

| Planner | Old | New | Faster | Slower | Same | Total |
|---------|-----|-----|--------|--------|------|-------|
| W | 1 | 2 | 161 | 61 | 30 | 252 |
| X | 1 | 2 | 295 | 126 | 0 | 421 |
| Y | 1 | 2 | 222 | 82 | 53 | 357 |
| Z | 1 | 2 | 84 | 121 | 30 | 235 |
| Z | 2 | 3 | 131 | 84 | 53 | 268 |
| Z | 3 | 4 | 115 | 92 | 21 | 228 |

Table 9: Improvements in execution speed across versions. The Faster column counts the number of cases in which the new version solved the problem faster; Slower specifies those cases in which the new version took longer to solve a given problem.

**Planner Assumption 2: Do Parameter Settings Matter to a Fair Comparison?**
In this planner set, only three have obvious, easily manipulable parameters: Blackbox, HSP and UCPOP. blackbox has an extensive set of parameters that control everything from how much trace information to print to the sequence of solver applications. HSP's function can be varied to include (or not) loop detection, change the search heuristic and vary the number paths to expand. For UCPOP, the user can change the strategies governing node orderings and flaw selection.

We did not run any experiments for this assumption because not all of the planners have parameters and because it is clear from the literature that the parameters do matter. Blackbox relies heavily on random restarts and trying alternative SAT solvers. In Kautz and Selman (1999), the authors of blackbox carefully study aspects of blackbox's design and demonstrate differential performance using different SAT solvers; they propose hypotheses for the performance differences and are working on better models of performance variation.

At the heart of HSP is heuristic search. Thus, its performance varies depending on the heuristics. Experiments with both HSP and FF (a planner that builds on some ideas from HSP) have shown the importance of heuristic selection in search space expansion, computation time and problem scale up (Haslum & Geffner, 2000; Hoffmann & Nebel, 2001).

As with HSP, heuristic search is critical to UCPOP's performance. A set of studies have explored alternative settings to the flaw selection heuristics employed by UCPOP (Joslin & Pollack, 1994; Srinivasan & Howe, 1995; Genevini & Schubert, 1996), producing dramatic improvements on some domains with some heuristics. As Pollack et al. (1997) confirmed, a good default strategy could be derived, but its performance was not the best under some circumstances.

Thus, because parameters can control fundamental aspects of algorithms, such as their search strategies, the role of parameters in comparisons cannot be easily dismissed.

**Planner Assumption 3: Are Time Cut-offs Unfair?** Planners often do not admit to failure. Instead, the planner stops when it has used the allotted time and not found a solution. So setting a time threshold is a requirement of any planner execution. In a comparison, one might always wonder whether enough time was allotted to be fair – perhaps the solution was almost found when execution was terminated.

To determine whether our cut-off of 30 minutes was fair, we examined the distribution of times for declared successes and failures[8]. Across the planners and the problem set, we found that the distributions were skewed (approximately log normal with long right tails) and that the planners were quick to declare success or failure, if they were going to do so. Table 10 shows the max, mean, median and standard deviation for success and failure times for each of the planners. The differences between mean and median indicate the distribution skew, as do the low standard deviations relative to the observed max times. The max time shows that on rare occasions the planners might make a decision within 2 minutes of our cut-off.

---

8. We separated the two because we usually observed a significant difference in the distributions of time to succeed and time to fail – about half the planners were quick to succeed and slow to fail, the other half reversed the relationship.

| Planner | Successes | | | | Failures | | | |
|---|---|---|---|---|---|---|---|---|
| | Max | Mean | Median | Sd | Max | Mean | Median | Sd |
| A | 667.9 | 34.0 | 1.3 | 98.7 | 1116.4 | 44.9 | 4.9 | 128.8 |
| B | 1608.5 | 38.5 | 0.5 | 182.8 | 1692.0 | 45.6 | 17.8 | 96.8 |
| C | 1455.4 | 89.9 | 1.6 | 244.6 | 1.4 | 0.4 | 0.13 | 0.4 |
| D | 481.0 | 17.8 | 1.1 | 77.4 | 713.6 | 26.3 | 1.1 | 122.6 |
| E | 1076 | 26.2 | 0.1 | 126.8 | 1622.8 | 286.9 | 260.6 | 189.1 |
| F | 1282.4 | 44.4 | 0.1 | 126.8 | 1188.4 | 22.3 | 0.2 | 104.8 |
| G | 1456.2 | 44.6 | 0.7 | 188.5 | 1196.5 | 43.8 | 16.7 | 78.5 |
| H | 657.7 | 29.58 | 1.4 | 80.6 | 1080.6 | 93.8 | 1.4 | 162.1 |
| I | 1713.8 | 115.4 | 0.2 | 303.1 | 50.6 | 5.1 | 4.9 | 6.3 |
| J | 1596.5 | 43.6 | 4.3 | 127.4 | 1796 | 11.0 | 11.0 | 57.9 |
| K | 1110.5 | 31.0 | 0.32 | 121.8 | 1298.8 | 27.7 | 12.1 | 65.2 |
| L | 1611.9 | 54.4 | 2.0 | 180.9 | 847.1 | 124.1 | 68.4 | 164.8 |
| M | 1675.3 | 53.4 | 1.45 | 196.5 | 1.6 | 0.9 | 0.8 | 0.4 |

Table 10: Max, mean, median and standard deviations (Sd) for the computation times to success and failure for each planner.

What this table does not show, but the observed distributions do show, is that very few values are greater than half of the time until the cut-off. Figures 2 and 3 display the distributions for planner F, which had means in the middle of the set of planners and quite typical distributions. Consequently, at least for these problems, any cut-off above 15 minutes (900 seconds) would not significantly change the results.

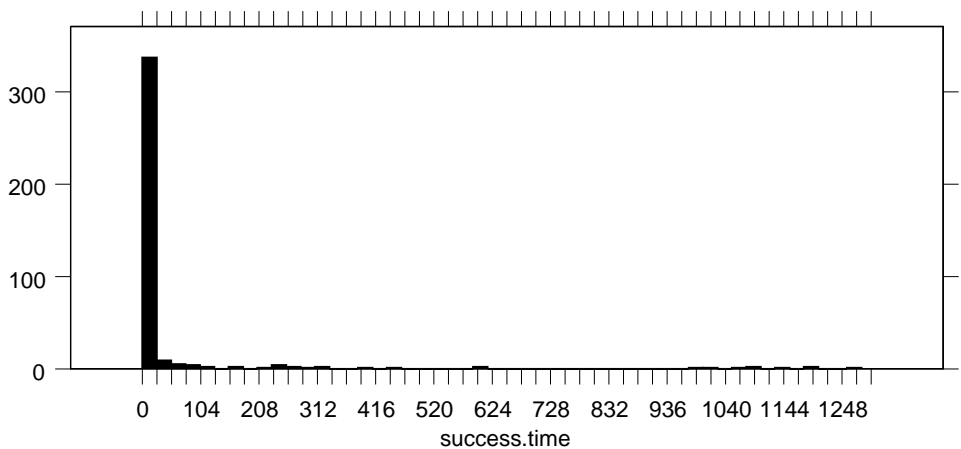

Figure 2: Histogram of times, in seconds, for planner F to succeed.

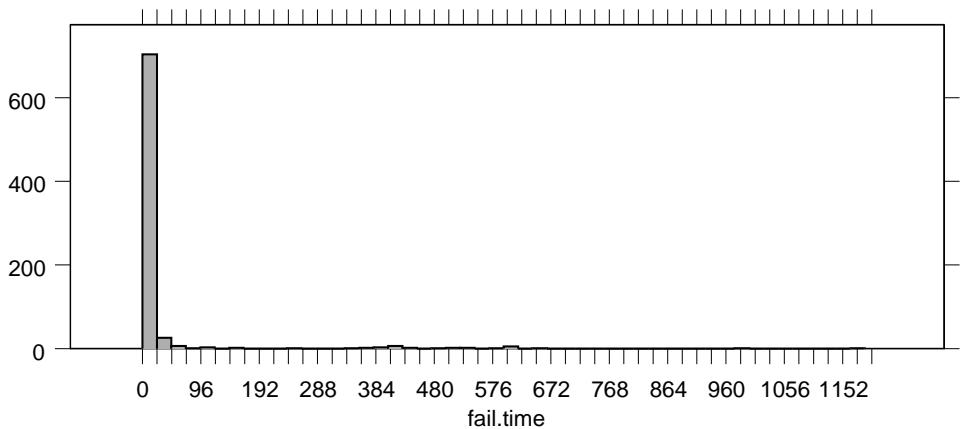

Figure 3: Histogram of times, in seconds, for planner F to fail.

## 3.4 Performance Metrics

Most comparisons emphasize the number of problems solved and the CPU time to completion as metrics. Often, the problems are organized in increasing difficulty to show scale-up. Comparing based on these metrics leaves a lot open to interpretation. For example, some planners are designed to find the optimal plan, as measured by number of steps in either a parallel or sequential plan. Consequently, these planners may require more computation. Thus, by ignoring plan quality, these planners may be unfairly judged. We also hypothesize that the hardware and software platform for the tests can vary the results. If a planner is developed for a machine with 1 GB of memory, then likely its performance will degrade with less. A key issue is whether the effect is more or less uniform across the set of planners.

In this section, we examine these two issues: execution platform and effect of plan quality.

**Metric Assumption 1: Does Performance Vary between Planners When Run on Different Hardware Platforms?**  Often when a planner is run at a competition or in someone else's lab, the hardware and software platforms differ from the platform used during development. Clearly, slowing down the processor speed should slow down planning, requiring higher cut-offs. Reduction in memory may well change the set of problems that can be solved or increase the processing time due to increased swapping. Changing the hardware configuration may change the way memory is cached and organized, favoring some planners' internal representations over others. Changing compilers could also affect the amount and type of optimizations in the code. The exact effects are probably unknown. The assumption is that such changes affect all planners more or less equally.

To test this, we ran the planners on a less powerful, lower memory machine and compared the results on the two platforms: the base Sun Ultrasparc 10/440 with 256mb of memory and Ultrasparc 1/170 with 128mb of memory. The operating system and compilers were the same versions for both machines. The same problems were run on both platforms. We followed much the same methodology as in the comparison of planner versions: comparing on both number of problems solved and time to solution. Table 11 shows the results as measured by problems solved, failed or timed-out for each planner on the two platforms.

| Planner | Platform | Solved | Failed | Timed-Out | $\chi^2$ | p | % Reduction |
|---------|----------|--------|--------|-----------|----------|-----|-------------|
| A | Ultra 1 | 94 | 383 | 27 | | | |
| | Ultra 10 | 95 | 389 | 20 | 1.09 | .58 | 1 |
| B | Ultra 1 | 121 | 346 | 37 | | | |
| | Ultra 10 | 121 | 353 | 30 | 0.80 | .67 | 0 |
| C | Ultra 1 | 354 | 7 | 143 | | | |
| | Ultra 10 | 367 | 7 | 130 | 0.85 | .65 | 4 |
| D | Ultra 1 | 218 | 59 | 227 | | | |
| | Ultra 10 | 217 | 59 | 228 | 0.01 | .998 | -.4 |
| E | Ultra 1 | 280 | 145 | 79 | | | |
| | Ultra 10 | 284 | 150 | 70 | 0.66 | .72 | 1 |
| F | Ultra 1 | 277 | 155 | 72 | | | |
| | Ultra 10 | 284 | 154 | 66 | 0.35 | .84 | 2 |
| G | Ultra 1 | 120 | 347 | 37 | | | |
| | Ultra 10 | 121 | 352 | 31 | 0.57 | .75 | 1 |
| H | Ultra 1 | 116 | 350 | 38 | | | |
| | Ultra 10 | 122 | 338 | 44 | 0.80 | .67 | 7 |
| I | Ultra 1 | 265 | 201 | 38 | | | |
| | Ultra 10 | 274 | 201 | 29 | 1.36 | .51 | 3 |
| J | Ultra 1 | 280 | 220 | 4 | | | |
| | Ultra 10 | 285 | 217 | 2 | 0.73 | .69 | 2 |
| K | Ultra 1 | 108 | 370 | 26 | | | |
| | Ultra 10 | 108 | 368 | 28 | 0.08 | .96 | 0 |
| L | Ultra 1 | 149 | 339 | 16 | | | |
| | Ultra 10 | 150 | 341 | 13 | 0.32 | .85 | 1 |
| M | Ultra 1 | 250 | 65 | 189 | | | |
| | Ultra 10 | 258 | 66 | 180 | 0.35 | .84 | 3 |

Table 11: Number of problems solved, failed and timed-out for each planner on the two hardware platforms. Last column is the percentage reduction in the number solved from the faster to slower platforms.

| Planner | Faster | | | Slower | | | Same | Total |
|---|---|---|---|---|---|---|---|---|
| | # | Mean $\Delta$ | Sd $\Delta$ | # | Mean $\Delta$ | Sd $\Delta$ | | |
| A | 92 | 5.18 | 30.76 | 1 | | | 1 | 94 |
| B | 120 | 4.02 | 10.01 | 0 | | | 1 | 121 |
| C | 294 | 31.89 | 101.71 | 60 | 0.29 | 0.14 | 0 | 354 |
| D | 177 | 11.02 | 82.82 | 39 | 0.23 | 0.14 | 1 | 217 |
| E | 275 | 2.68 | 12.27 | 1 | | | 4 | 280 |
| F | 271 | 14.86 | 72.44 | 0 | | | 6 | 277 |
| G | 117 | 5.02 | 17.17 | 1 | | | 2 | 120 |
| H | 115 | 6.86 | 25.24 | 0 | | | 1 | 116 |
| I | 261 | 25.73 | 119.97 | 0 | | | 4 | 265 |
| J | 280 | 42.24 | 138.16 | 0 | | | 0 | 280 |
| K | 107 | 15.26 | 75.42 | 0 | | | 1 | 108 |
| L | 148 | 16.81 | 98.54 | 1 | | | 0 | 149 |
| M | 194 | 32.72 | 139.73 | 56 | 0.30 | 0.18 | 0 | 250 |

Table 12: Improvements in execution speed moving from slower to faster platform. Counts only problems that were solved on both platforms. For faster and slower, the mean and standard deviation (Sd) of difference is also provided.

As before, we also looked at change in time to solution. Table 12 shows how the time to solution changes for each planner. Not surprisingly, faster processor and more memory nearly always lead to better performance. Somewhat surprisingly, the difference is far less than the doubling that might be expected; the mean differences are much less than the mean times on the faster processor (see Table 10 for the mean solution times).

Also, the effect seems to vary between the planners. Based on the counts, the Lisp-based planners appear to be less susceptible to this trend (the only ones that sometimes were faster on the slower platform). However, the advantages are very small, affecting primarily the smaller problems. We think that this effect is due to the need to load in a Lisp image at startup from a centralized server; thus, computation time for small problems will be dominated by any network delay. Older versions of planners appear to be less sensitive to the switch in platform.

In this study, the platforms make little difference to the results, despite a more than doubling of processor speed and doubling of memory. However, the two platforms are underpowered when compared to the development platforms for some of the planners. We chose these platforms because they differed in only a few characteristics (processor speed and memory amount) and because we had access to 20 identically configured machines. To really observe a difference, $1\text{GB}^9$ of memory or more may be needed.

Recent trends in planning technology have exploited cheap memory: translations to propositional representations, compilation of the problems and built-in caching and memory management techniques. Thus, some planners are designed to trade-off memory for time;

---

9. We propose this figure because it is the amount requested by some of the participants in the AIPS 2000 planning competition.

these planners will understandably be affected by memory limitations for some problems. Given the results of this study, we considered performing a more careful study of memory by artificially limiting memory for the planners but did not do so because we did not have access to enough sufficiently large machines to likely make a difference and because we could not devise a scheme for fairly doing so across all the planners (which are implemented in different languages and require different software run-time environments).

Another important factor may be memory architecture/management. Some planners include their own memory managers, which map better to some hardware platforms than to others (e.g., HSP uses a linear organization that appears to fit well with Intel's memory architecture).

**Metric Assumption 2: Do the Number of Plan Steps Vary?** Several researchers have examined the issue of measuring plan quality and directing planning based on it, e.g., (Perez, 1995; Estlin & Mooney, 1997; Rabideau, Englehardt, & Chien, 2000). The number of steps in a plan is a rather weak measure of plan quality, but so far, it is the only one that has been widely used for primitive-action planning.

We expect that some planners sacrifice quality (as measured by plan length) for speed. Thus, ignoring even this measure of plan quality may be unfair to some planners. To check whether this appears to be a factor in our problem set, we counted the plan length in the plans returned in output and compared the lengths across the planners. Because not all of the planners construct parallel plans, we adopted the most general definition: sequential plan length. We then compared the plan lengths returned by each planner on every successfully solved problem.

We found that 11% of the problems were solved by only one planner (not necessarily the same one). The planners found equal length solutions for 62% of those that remained (493 problems). We calculated the standard deviation (SD) of plan length for solutions to each problem and then analyzed the SDs. We found that the minimum observed SD was 0.30, the maximum was 63.30, the mean was 2.43 and the standard deviation was 5.45. Thirteen cases showed SDs higher than 20. Obviously, these cases involved fairly long plans (up to 165 steps); the cases were for problems from the logistics and gripper domains.

To check whether some planners favored minimal lengths, we counted the number of cases in which each planner found the shortest length plan (ties were attributed to all planners) when there was some variance in plan length. Table 13 lists the results. Most planners find the shortest length plans on about one third of these problems. Planner F was designed to optimize plan length, which shows in the results. With one exception, the older planners rarely find the shortest plans.

## 4. Interpretation of Results and Recommendations

The previous section presented our summarization and analysis of the planner runs. In this section, we reflect on what those results mean for empirical comparison of planners; we summarize the results and recommend some partial solutions. It is not possible to guarantee fairness and we propose no magic formula for performing evaluations, but the state of the practice in general can certainly be improved. We propose three general recommendations and 12 recommendations targeted to specific assumptions.

| Planner | Count |
|---------|------:|
| A | 178 |
| B | 169 |
| C | 0 |
| D | 161 |
| E | 5 |
| F | 319 |
| G | 171 |
| H | 176 |
| I | 222 |
| J | 0 |
| K | 159 |
| L | 151 |
| M | 283 |

Table 13: Number of plans on which each planner found the shortest plan. The data only include problems for which different length plans were found.

Many of the targeted recommendations amount to requesting problem and planner developers to be more precise about the requirements for and expectations of their contributions. Because the planners are extremely complex and time consuming to build, the documentation may be inadequate to determine how a subsequent version differs from the previous or under what conditions (e.g., parameter settings, problem types) the planner can be fairly compared. With the current positive trend in making planners available, it behooves the developer to include such information in the distribution of the system.

The most sweeping recommendation is to shift the research focus away from developing *the best* general-purpose planner. Even in the competitions, some of the planners identified as superior have been ones designed for specific classes of problems, e.g., FF and IPP. The competitions have done a great job of exciting interest and encouraging the development and public availability of planners that incorporate the same representation.

However, to advance the research, the most informative comparative evaluations are those designed for a specific purpose – to test some hypothesis or prediction about the performance of a planner[10]. An experimental hypothesis focuses the analysis and often leads naturally to justified design decisions about the experiment itself. For example, Hoffmann and Nebel, the authors of the Fast-Forward (FF) system, state in the introduction to their JAIR paper that FF's development was motivated by a specific set of the benchmark domains; because the system is heuristic, they designed the heuristics to fit the expectations/needs of those domains (Hoffmann & Nebel, 2001). Additionally, in part of their evaluation, they compare to a specific system on which their own system had commonalities and point out the various advantages or disadvantages of their design decisions on specific problems. Follow-up work or researchers comparing their own systems to FF now have a well-defined starting point for any comparison.

---

10. Paul Cohen has advocated such an experimental methodology for all of artificial intelligence based on hypotheses, predictions and models in considerable detail; see Cohen (1991, 1995).

> **Recommendation 1:** Experiments should be driven by hypotheses. Researchers should precisely articulate in advance of the experiments their expectations about how their new planner or augmentations to an existing planner add to the state of the art. These expectations should in turn justify the selection of problems, other planners and metrics that form the core of the comparative evaluation.

A general issue is whether the results are accurate. We reported the results as they are output by the planners. If a planner stated in its output that it had been successful, we took it at face value. However, by examining some of the output, we determined that some claims of successful solution were erroneous – the proposed solution would not work. The only way to ensure that the output is correct is with a solution checker. Drew McDermott used a solution checker in the AIPS98 competition. However, the planners do not all provide output in a compatible format with his checker. Thus, another concern with any comparative evaluation is that the output needs to be cross-checked. Because we are not declaring a winner (i.e., that some planner exhibited superior performance), we do not think that the lack of a solution checker casts serious doubt on our results. For the most part, we have only been concerned with factors that cause the observed success rates to change.

> **Recommendation 2:** Just as input has been standardized with PDDL, output should be standardized, at least in the format of returned plans.

Another general issue is whether the benchmark sets are representative of the space of interesting planning problems. We did not test this directly (in fact, we are not sure how one could do so), but the clustering of results and observations by others in the planning community suggest that the set is biased toward logistics problems. Additionally, many of the problems are getting dated and no longer distinguish performance. Some researchers have begun to more formally analyze the problem set, either in service of building improved planners (e.g., Hoffmann & Nebel, 2001) or to better understand planning problems. For example, in the related area of scheduling, our group has identified distinctive patterns in the topology of search spaces for different types of classical scheduling problems and has related the topology to performance of algorithms (Watson, Beck, Barbulescu, Whitley, & Howe, 2001). Within planning, Hoffmann has examined the topology of local search spaces in some of the small problems in the benchmark collection and found a simple structure with respect to some well-known relaxations (Hoffmann, 2001). Additionally, he has worked out a partial taxonomy, based on three characteristics, for the analyzed domains. Helmert has analyzed the computational complexity of a subclass of the benchmarks, transportation problems, and has identified key features that affect the difficulty of such problems (Helmert, 2001).

> **Recommendation 3:** The benchmark problem sets should themselves be evaluated and over-hauled. Problems that can be easily solved should be removed. Researchers should study the benchmark problems/domains to classify them into problem types and key characteristics. Developers should contribute application problems and realistic versions of them to the evolving set.

The remainder of this section describes other recommendations for improving the state of the art in planner comparisons.

**Problem Assumption 1: Are General Purpose Planners Biased Toward Particular Problems/Domains?** The set of problems on which a planner was developed can have a strong effect on the performance of the planner. This can be either the effect of unintentional over-specialization or the result of a concerted effort on the part of the developers to optimize their system to solve a specific problem. With one exception, every planner fared better on the tailored subset of problems (training set). Consequently, we must conclude that the choice of a subset of problems may well affect the outcome of any comparison.

A fair planner comparison must account for likely biases in the problem set. Good performance on a certain class of problems does not imply good performance in general. A large performance differential for planners with a targeted problem domain (i.e., do well on their focus problems and poorly on others) may well indicate that the developers have succeeded in optimizing the performance of their planner.

> **Recommendation 4:** Problem sets should be constructed to highlight the designers' expectations about superior performance for their planner, and they should be specific about this selection criteria.

On the other hand, if the goal is to demonstrate across the board performance, then our results at randomly selecting domains suggests that biases can be mitigated.

> **Recommendation 5:** If highlighting performance on "general" problems is the goal, then the problem set should be selected randomly from the benchmark domains.

**Problem Assumption 2: How Do Syntactic Representation Differences Affect Performance?** Many studies, including this, have shown that planners may be sensitive to representational features. Just because representations can be translated automatically does not mean that performance will be unaffected. Just because an algorithm should theoretically be insensitive to a factor does not mean that in practice it is. All of the planners showed some sensitivity to permuted problems, and the degree of sensitivity varied. This outcome suggests that translators and even minor variations on problem descriptions impact outcome and should be used with care, especially when the sensitivity is not the focus of the study and some other planner is more vulnerable to the effect.

> **Recommendation 6:** Representation translators should be avoided by using native versions of problems and testing multiple versions of problems if necessary.

With many planner developers participating in the AIPS competitions, this should become less of an issue.

More importantly, researchers should be explicitly testing the effect of alternative phrasings of planning problems to determine the sensitivity of performance and to separate the effects of advice/tuning from the essence of the problem.

**Recommendation 7:** Studies should consider the role of minor syntactic variations in performance and include permuted problems (i.e., initial conditions, goals, preconditions and actions) in their problem sets because they can demonstrate robustness, provide an opportunity for learning and protect developers from accidentally over-fitting their algorithm to the set of test problems.

**Problem Assumption 3: Does Performance Depend on PDDL Requirements Features?** The planners did not perform quite as advertised or expected given some problem features. This discrepancy could have many possible causes: problems incorrectly specified, planners with less sensitivity than thought, solutions not being correct, etc. For example, many of the problems in the benchmark set were not designed for the competitions or even intended to be widely used and so may not have been specified carefully enough.

**Recommendation 8:** When problems are contributed to the benchmark set, developers should verify that the requirements stated in the description of each problem correctly reflect the subset of features needed. Planner evaluators should then use only those problems that match a planner's capabilities.

Depending on the cause, the results can be skewed, e.g., a planner may be unfairly maligned for being unable to solve a problem that it was specifically designed not to solve. The above recommendation addresses gaps in the specification of the problem set, but some mismatches between the capabilities specifiable in PDDL and those that planners possess remain.

**Recommendation 9:** Planner developers should develop a vocabulary for their planner's capabilities, as in the PDDL flags, and specify the expected capabilities in the planner's distribution.

**Planner Assumption 1: Is the Latest Version the Best?** Our results suggest that new versions run faster, but often do not solve more problems. Thus, the newest version may not represent the "best" (depending on your definition) performance for the class of planner. Some competitions in other fields, e.g., the automatic theorem proving community, require the previous year's best performer to compete as well; this has the advantage of establishing a baseline of performance as well as allowing a comparison to how the focus may shift over time.

**Recommendation 10:** If the primary evaluation metric is speed, then a newer version may be the best competition. If it is number of problems solved or if one wishes to establish what progress has been made, then it may be worth running against an older version as well. If recommendation 9 has been followed, then evaluators should select a version based on this guidance.

**Planner Assumption 2: What Are the Effect of Parameter Settings?** Performance of some planners does vary with the parameter settings. Unfortunately, it often is difficult to figure out how to set the parameters properly, and changing settings makes it difficult to compare results across experiments. Generally, this is not an issue because the developers and other users tend to rely on the default parameter settings. Unfortunately, sometimes the developers exploit alternative settings in their own experiments, complicating later comparison.

> **Recommendation 11:** If a planner includes parameters, the developer should guide users in their settings. If they do not, then the default settings should be used by both the developers and others in experiments to facilitate comparison.

**Planner Assumption 3: Are Time Cut-offs Unfair?** We found little benefit from increasing time cut-offs beyond 15 minutes for our problems.

> **Recommendation 12:** If total computation time is a bottleneck, then run the problems in separate batches, incrementally increasing the time cut-off between runs and including only unresolved problems in subsequent runs. When no additional problems are solved in a run, stop.

**Metric Assumption 1: Do Alternative Platforms Lead to Different Performance?** In our experiments, performance did not vary as much as we expected. This result suggests that researchers in general are not developing for specific hardware/software configurations, but recent trends suggest otherwise, at least with regards to memory. Again, because these systems are research prototypes, it behooves the developer to be clear about his/her expectations and anyone subsequently using the system to accommodate those requests in their studies.

> **Recommendation 13:** As with other factors in planner design, researchers must clearly state the hardware/software requirements for their planners, if the design is based on platform assumptions. Additionally, a careful study of memory versus time trade-offs should be undertaken, given the recent trends in memory exploitation.

**Metric Assumption 2: Do the Number of Plan Steps Vary?** They certainly can. If one neglects quality measures, then some planners are being penalized in efforts to declare a best planner.

> **Recommendation 14:** To expedite generalizing across studies, reports should describe performance in terms of what was solved (how many of what types), how much time was required and what were the quality of the solutions. Trade-offs should be reported, when possible, e.g., 12% increase in computation time for 30% decrease in plan length. Additionally, if the design goal was to find an optimal solution, compare to other planners with that as their design goal.

Good metrics of plan quality are sorely needed. The latest specification of the PDDL specification supports the definition of problem-specific metrics (Fox & Long, 2002); these metrics indicate whether total-time (a new concept supported by specification of action durations) or specified functions should be minimized or maximized. This addition is an excellent start, but general metrics other than just plan-length and total-time are also needed to expedite comparisons across problems.

> **Recommendation 15:** Developing good metrics is a valuable research contribution. Researchers should consider it a worthwhile project, conference organizers and reviewers should encourage papers on the topic, and planner developers should implement their planners to be responsive to new quality metrics (i.e., support tunable heuristics or evaluation criteria).

## 5. Conclusions

Fair evaluation and comparison of planners is hard. Many apparently benign factors exert significant effects on performance. Superior performance of one planner over another on a problem that neither was intentionally designed to solve may be explained by minor representational features. However, comparative analysis on general problems is of practical importance as it is not practical to create a specialized solution to every problem.

We have analyzed the effects of experiment design decisions in empirical comparison of planners and made some recommendations for ameliorating the effects of these decisions. Most of the recommendations are common sense suggestions for improving the current methodology.

To expand beyond the current methodology will require at least two substantive changes. First, the field needs to question whether we should be trying to show performance on planning problems in general. A shift from general comparisons to focused comparisons (on problem class or mechanism or on hypothesis testing) could produce significant advances in our understanding of planning.

Second, the benchmark problem sets require attention. Many of the problems should be discarded because they are too simple to show much. The domains are far removed from real applications. It may be time to revisit testbeds. For example, several researchers in robotics have constructed an interactive testbed for comparing motion planning algorithms (Piccinocchi, Ceccarelli, Piloni, & Bicchi, 1997). The testbed consists of a user interface for defining new problems, a collection of well-known algorithms and a simulator for testing algorithms on specific problems. Thus, the user can design his/her own problems and compare performance of various algorithms (including their own) on them via a web site. Such a testbed affords several advantages over the current paradigm of static benchmark problems and developer conducted comparisons, in particular, replicability and extendability of the test set. Alternatively, challenging problem sets can be developed by modifying deployed applications (Wilkins & desJardins, 2001; Engelhardt, Chien, Barrett, Willis, & Wilklow, 2001).

In recent years, the planning community has significantly improved the size of planning problems that can be solved in reasonable time and has advanced the state of the art in empirical comparison of our systems. To interpret the results of empirical comparisons and understand how they should motivate further development in planning, the community needs to understand the effects of the empirical methodology itself. The purpose of this paper is to further that understanding and initiate a dialogue about the methodology that should be used.

## Acknowledgments

This research was partially supported by a Career award from the National Science Foundation IRI-9624058 and by a grant from Air Force Office of Scientific Research F49620-00-1-0144. The U.S. Government is authorized to reproduce and distribute reprints for Governmental purposes notwithstanding any copyright notation thereon. We are most grateful to the reviewers for the careful reading of and well-considered comments on the submitted version; we hope we have done justice to your suggestions.